\documentclass{article}

\PassOptionsToPackage{numbers, compress}{natbib}
\usepackage[preprint]{nips_2020}

\usepackage[utf8]{inputenc} 
\usepackage[T1]{fontenc}    
\usepackage{hyperref}       
\usepackage{url}            
\usepackage{booktabs}       
\usepackage{amsfonts}       
\usepackage{nicefrac}       
\usepackage{microtype}      

\usepackage{hyperref}
\usepackage{url}

\usepackage{amsmath}
\usepackage{verbatim}
\usepackage{amssymb}
\usepackage{amsthm}
\usepackage{algorithm}
\usepackage{algorithmic}
\usepackage{graphicx}
\usepackage{natbib}
\usepackage{booktabs}
\usepackage{subfigure}
\usepackage{epstopdf}
\usepackage{lipsum}
\usepackage{color}

\usepackage{multicol}

\definecolor{shadecolor}{gray}{0.95}

\usepackage{graphicx} 
\usepackage{subfigure}

\usepackage{natbib}

\usepackage{algorithm}
\usepackage{algorithmic}

\usepackage{hyperref}
\usepackage{amsthm}
\usepackage{amssymb}
\usepackage{multicol}

\usepackage{amsmath}

\usepackage{bbm,amssymb,amsmath,amsfonts,graphicx,ifthen,twoopt,algorithm,algorithmic,color}

\def\E{\mathbb{E}}

\newtheorem{thm}{Theorem}
\newtheorem{lem}{Lemma}

\title{Contextual Bandit with Missing Rewards}

\author{Djallel Bouneffouf, Sohini Upadhyay and Yasaman Khazaeni \\
  IBM Research AI\\
  \texttt{\{firstname.lastname\}@ibm.com} \\
}
\newtheorem{definition}{Definition}

\begin{document}

\maketitle

\begin{abstract}
We consider a novel variant of the contextual bandit problem (i.e., the multi-armed bandit with side-information, or context, available to a decision-maker) where the reward associated with each context-based decision may not always be observed (``missing rewards'').   This new problem is motivated by certain online settings including clinical trial and ad recommendation applications. In order to address the missing rewards setting, we propose to combine the standard contextual bandit approach with an unsupervised learning mechanism such as clustering. Unlike standard contextual bandit methods, by leveraging clustering to estimate missing reward, we are  able to learn from each incoming event, even those with missing rewards. Promising empirical  results are obtained on several real-life datasets.
\end{abstract}

\section{Introduction}
Sequential decison making is a common problem in many practical applications, broadly encompassing situations in which an agent must choose the best action to perform at each iteration while maximizing cumulative reward over some period of time \cite{BouneffoufBG13,ChoromanskaCKLR19,RiemerKBF19,LinC0RR20,lin2020online,lin2020unified,NoothigattuBMCM19,lin2019split,lin2020story}.
One of the key challenges in sequential decision making is to achieve a good trade-off between the exploration of new actions and the exploitation of known actions.  This exploration vs exploitation trade-off in sequential decision making is often formulated as the {\em multi-armed bandit (MAB)} problem. In the MAB problem setting, given a set of bandit ``arms'' (actions),  each associated with a fixed but unknown reward probability distribution ~\cite {surveyDB,LR85,UCB,Bouneffouf0SW19,LinBCR18,DB2019,BalakrishnanBMR19ibm,BouneffoufLUFA14,RLbd2018,balakrishnan2020constrained,BouneffoufRCF17},  the agent selects an arm to play at each iteration, and receives a reward,  drawn according to the selected arm's distribution, independently from the previous actions. 

A particularly useful version of MAB is the {\em contextual multi-armed bandit (CMAB)}, or simply the {\em contextual bandit} problem, where at each iteration, the agent observes a $N$-dimensional {\em context}, or {\em feature vector} prior to choosing an arm \cite{AuerC98,AuerCFS02,BalakrishnanBMR18,BouneffoufBG12}.
Over time, the goal is to learn the relationship between the context vectors and rewards, in order to make better action choices given the  context \cite{AgrawalG13}. Common sequential decision making problems with side information (context) that utilize the contextual bandit approach range from clinical trials \cite{villar2015multi} to recommender systems \cite{MaryGP15,Bouneffouf16,aaai0G20}, where the patient's information (medical history, etc.) or an online user profile provide a context for making better decisions about which treatment to propose or ad to show. The reward reflects the outcome of the selected action, such as success or failure of a particular treatment option, or whether an ad is clicked or not.

In this paper we consider a new problem setting referred to as {\em contextual bandit with missing rewards}, where the agent can always observe the context but may not always observe the reward.  
This setting is motivated by several real-life applications where the reward associated with a selected action can be missing, or unobservable by the agent, for various reasons. For instance, in medical decision making settings, a doctor can decide on a specific treatment option for a patient, but the patient may not come back for follow-up appointments; though the reward feedback regarding the treatment success is missing, the context, in this case the patient's medical record, is still available  and can be potentially used to learn more about the patient's population. Missing rewards can also occur in information retrieval or online search settings where a user enters a search request, but, for various reasons, may not click on any of the suggested website links, and thus the reward feedback about those choices is missing. Yet another example is in online advertisement, where a user clicking on a proposed ad represents a positive reward, but the absence of a click can be negative reward (the user did not like the ad), or can be a consequence of a bug or connection loss.

The contextual bandit with missing rewards framework proposed here aims to capture the situations described above, and provide an approach to exploit all context information for future decision making, even if some rewards are missing. More specifically, we will combine unsupervised online clustering with the standard contextual bandit. Online clustering allows us to learn representations of all the context vectors, with or without the observed rewards. Utilizing the contextual bandit on top of clustering makes use of the reward information when it is available. We demonstrate on several real-life datasets that this  approach consistently outperforms the standard contextual bandit approach when rewards are missing.

\section{Related Work}
\label{sec:related}
The multi-armed bandit problem provides a solution to the exploration versus exploitation trade-off \cite {AllesiardoFB14,dj2020,Sohini2019}. This problem has been extensively studied. Optimal solutions have been provided using a stochastic formulation ~\cite {LR85,UCB,BouneffoufF16}, a Bayesian formulation ~\cite {T33}, and an adversarial formulation ~\cite{AuerC98,AuerCFS02}. However, these approaches do not take into account the relationship between context and reward, potentially inhibiting overall performance.
In LINUCB ~\cite{Li2010,ChuLRS11} and in Contextual Thompson Sampling (CTS)~\cite{AgrawalG13}, the authors assume a linear dependency between the expected reward of an action and its context; the representation space is modeled using a set of linear predictors. However, these algorithms assume that the bandit can observe the reward at each iteration, which is not the case in many practical applications, including those discussed earlier in this paper. Authors in \cite{bartok2014partial} considered a kind of incomplete feedback called "Partial  Monitoring (PM)", developing a general framework for sequential decision making problems with incomplete feedback. The framework allows the learner to retrieve the expected value of actions through an analysis of the feedback matrix when possible, assuming both are known to the learner.

In \cite{bouneffouf2020online}, authors study a variant of the stochastic multi-armed  bandit (MAB) problem in which the context are corrupted. The new problem is motivated by certain online settings including clinical trial and ad recommendation applications. In order to address the corrupted-context setting, the author propose to combine the standard contextual bandit approach with a classical multi-armed bandit mechanism. Unlike standard contextual bandit methods, they were able to learn from all iteration, even those with corrupted context, by improving the computing of the expectation for each arm. Promising empirical results are obtained on several real-life datasets. 

In this paper we focus on handling incomplete feedback in the bandit problem setting more generally, without assuming the existence of a systematic corruption process. Our work is somewhat comparable to online semi-supervised learning \cite{Yver2009, ororbia2015online}, a field of machine learning that studies learning from both labeled and unlabeled examples in an online setting. However, in online semi-supervised learning, the true label is available at each iteration, whereas in the contextual bandit with missing rewards, only bandit feedback is available, and the true label, or best action, is unknown.


\section{Problem Setting} 

{ Algorithm \ref{alg:CBP1} presents at a high-level  the contextual bandit setting, where $x_t\in C$ (we will assume here  $C = \mathbf{R}^N$) is a vector  describing the context  at time $t$,  $r_{t,i} \in [0,1]$ is the  reward of the action $i$ at time $t$, and $r_t \in [0,1]^K$ denotes a vector of  rewards for all arms at time $t$. Also, $D_{c,r}$ denotes a joint probability distribution over $(x,r)$, $A$ denotes a set of $K$ actions, $A = \{1,...,K\}$, and $\pi: C \rightarrow A$ denotes a policy.
We operate under the linear realizability assumption; that is, there exists an unknown weight vector $\theta^* \in R$ with $ ||\theta^*||\leq 1$ so that,

\begin{equation*}
\forall k, t: \; \E[r_k(t) \vert x_t] = \theta_k^\top x_t  
+ n_t .\end{equation*}

where $\theta_k \in \mathbb{R}^d$ is an unknown coefficient vector associated with the arm $k$ which needs to be learned from data. Hence, we assume that the $r_{t,k}$ are independent random variables with expectation $x^\top \theta^*+ n_t$. with $n_t$ some measurement noise.
We also assume here that, the measurement noise $n_t$ is independent of everything and is $\sigma$-sub-Gaussian for some $\sigma >0$, i.e., $E[e^{\phi\, n_t} ] \leq exp(\frac{\phi^2 \sigma^2}{2})$ for all $ \phi \in R$.
\begin{definition}[Cumulative regret]
{The regret of an algorithm accumulated during $T$ iterations is given as:
\begin{equation*}
R(T) =\sum ^{T}_{t=1} r_{t,k^*(t)} - \sum^{T}_{t=1} r_{t,k(t)}.
\end{equation*}}
\end{definition}
 where $k^*(t)= \text{argmax}_k x_{t}^\top \theta^*$is the best action at step $t$ according to $\theta^*$.}

\begin{algorithm}[H]
	\caption{ Contextual Bandit }
	\label{alg:CBP1}
	\begin{algorithmic}[1]
		\STATE {\bfseries }\textbf{Repeat}
		\STATE {\bfseries } $(x_t,r_t)$ is drawn according to $D_{x,r}$
		\STATE {\bfseries }$x_t$ is revealed to the player
		\STATE {\bfseries } The player chooses an action $k =\pi_t(x_t)$
				\STATE {\bfseries } The reward $r_t$
		\STATE {\bfseries } The player updates its policy $\pi_t$
			\STATE {\bfseries } $t=t+1$
		\STATE {\bfseries }\textbf{Until} t=T
	\end{algorithmic}
\end{algorithm}

\section{LINUCB with Missing Rewards (MLINUCB)}
\label{sec:banditDL}
One solution for the contextual bandit is the LINUCB algorithm~\cite{13} where the key idea is to apply online ridge regression to incoming data to obtain an estimate of the coefficients $\theta_k$. 
In order to make use of the context even in the absence of the corresponding reward, we propose to use an  unsupervised learning approach; specifically, we use an online clustering step to retrieve missing rewards from available rewards with similar contexts. At each time step, the context vectors $x(t)$ are clustered into $N$ clusters where $N$ is selected \emph{a-priori}.

We adapt the LINUCB algorithm for our setting, proposing to use a clustering step for imputing the reward data when missing. At each time step, we perform a clustering step on the context vectors where the total number of clusters $N$ is a hyperparameter. For each cluster $j$ we define the average reward for each arm as below:
\begin{equation}
\overline{r}_j=\frac{\sum_{\tau=1}^{n_j} r_{\tau}}{n_j}
\label{waverage}
\end{equation}
Assuming $d_j=dist(x_t,\gamma_j)$ is the metric used for clustering where $\gamma_j$ is the $j^{th}$ cluster centroid and $n_j$ is the number of data points in cluster $j$, we choose the $m$ smallest $d_j$ as the closest clusters to $x_t$ and compute a weighted average of the average cluster rewards as formulated below: \begin{equation}g(x_t)=\frac{\sum_{j=1}^{m}\frac{\overline{r}_j}{d_j}}{\sum_{j=1}^{m}\frac{1}{d_j}}\end{equation}
When $r_t$ is missing we assign
$r_t=g(x_t)$. Note that if $m=1$, $g(x_t)$ is simply the average rewards of all the points within the cluster that $x_t$ belongs to.  
 
\begin{algorithm}[H]
\caption{MLINUCB}
 \label{alg:LINUCB}
 \begin{algorithmic}[1]
   \STATE {\bfseries Input:} value for $\alpha$, $b_0$, $\textbf{A}_0$, $N$, $m$
   \FOR{t=1 {\bfseries to} T} 
   \STATE cluster \{$x_1$, ... , $x_t$\} into $N$ clusters
   \FOR{all $k \in K$}
   \STATE $\theta_k \leftarrow$ \textbf{A}$_{k_t}^{-1}*b_{k_t}$  
   \STATE $p_{t,k} \leftarrow \theta^{\top}_k x_{t} +
   \alpha \sqrt{x^{\top}_{t}
   \textbf{A}_{k_t}^{-1} x_{t}}$
    \ENDFOR
   \STATE Choose arm $k_t = \text{argmax}_{k\in K}  p_{t,k}$,
   and observe real-valued payoff $r_t$
   \IF{ $r_t$ available}
   \STATE retrieve $r_t$ from data
    \ELSE 
   \STATE $r_t \leftarrow g(x_t)$.
        \ENDIF 
   \STATE \textbf{A}$_{k_t} \leftarrow$ \textbf{A}$_{k_t} + x_{t,k_t} x^{\top}_{t,k_t}$
   \STATE $b_{k_t} \leftarrow b_{k_t} + r_t x_{t,k_t}$ 
   \ENDFOR
   \end{algorithmic}
\end{algorithm}

We now upper bound the regret of MLINUCB. Note that the general CBP setting  \cite{abbasi2011improved} takes one context per arm instead for our setting of the one context share by actions. To upper bound our algorithm for the general CBP setting, we simply cast our setting as theirs by the following steps.
We simply choose a global vector $\theta$ as the concatenation of the $K$ vectors, so $\theta =[\theta_{1},...,\theta_{K}]$. We define a context $x_{t,k}$ per action with $x_t$, where $x_{t,k} =[...0,x_{t}^{\top},0,... ]^{\top}$ and $x_t$ being the $k$-th vector within the concatenation. All $A_t$,$r_t$, $b_t$ can be similarly defined from $A_{k(t)}$, $r_{k(t)}$, $b_{k(t)}$.
\begin{thm} \label{thm:hlinucb}
With probability $1-\delta$, where $0 < \delta < 1$, the upper bound on the R(T) for the MLINUCB in the contextual bandit problem, $K$ arms and $d$ features (context size) is given as follows:
\begin{eqnarray}
R(T)\leq \sigma (\sqrt{d \;log (\frac{det(A_{T})^{1/2} }{\delta\;det(\mathbf{S})^{1/2}} )}+\nonumber \\ \frac{||\theta||}{\sqrt{\phi}})\sqrt{18\; T log(\frac{det(A_{T})}{det(\mathbf{S})})} 
\end{eqnarray}
with $||x_t||_2 \leq L$, with $\mathbf{S}=\mathbf{I}+ \sum_{t \in s} x_tx_t^\top $ with $s \subset T$ contains the contexts with missing rewards and $\phi \in R$
\end{thm}
Theorem \ref{thm:hlinucb} shows that MLINUCB has better upper bound compared to the LINUCB \cite{abbasi2011improved}, where in LINUCB upper bound has $log(det(A_{T}))$ under the square root where we have $log(\frac{det(A_{T})}{det(\mathbf{H})})$. We can see that the upper bound depends on $H$, so more context with missing rewards better is the bound.

\section{Regret analysis of BILINUCB}
We now upper bound the regret of BILINUCB. General Contextual Bandit Problem (CBP) setting \cite{abbasi2011improved} assumes one context per arm instead for BILINUCB setting with same context shared across arms. To upper bound regret of BILINUCB we cast our setting as general CBP setting in the following way.
We choose a global vector $\theta$ as the concatenation of the $K$ vectors, so $\theta =[\theta_{1},...,\theta_{K}]$. Next define a context $x_{t,k}$ per arm as $x_{t,k} =[...,0,x_{t}^{\top},0,... ]^{\top}$ with $x_t$ being the $k$-th vector within the concatenation. Let $\mathbf{S_T} = \mathcal{I}_D + \sum_{t \in s} x_t x_t^\top $, where $s \subset \{1,\ldots,T\}$ contains the contexts with missing rewards up to step $T$, and let $\mathbf{A}_T = \mathbf{S_T} + \sum_{t \not\in s} x_t x_t^\top $. We have the following theorem regarding the regret bound up to step $T$.

Theorem 1 of the main text shows that BILINUCB has better upper bound compared to the LINUCB \cite{abbasi2011improved}, where in LINUCB upper bound has $\log(\det(A_{t}))$ under the square root where we have $\log(\frac{\det(A_{t})}{\det(\mathbf{S}_T)})$. 
The matrix $\mathbf{S}_T$ is the sum of identity matrix $\mathcal{I}_D$ and covariance matrix $\Sigma_s= \sum_{t \in s} x_t x_t^\top $ constructed using the contexts with missing reward. Both $\mathcal{I}_D$ and $\Sigma_s$ are real symmetric and hence Hermitian matrices. Further, $\Sigma_s$ is positive semi-definite as a covariance matrix.
Since all the eigenvalues of $\mathcal{I}_D$ equal $1$ and since all the eigenvalues of $\Sigma_s$ are non-negative, by Weyl's inequality in matrix theory for perturbation of Hermitian matrices, the eigenvalues of $\mathbf{S}_T$ are lower bounded by $1$. Hence $\det(\mathbf{S}_T)$ which is the product of the eigenvalues of $\mathbf{S}_T$ is lower bounded by $1$. Hence, BILINUCB which involves the term $\frac{\det (\mathbf{A}_T)}{\det (\mathbf{S}_T)}$ has a provably better guarantee than LINUCB which involves only the term $\det(\mathbf{A}_T)$ (without $\det(\mathbf{S}_T)$).
\subsection{Proof of Theorem 1}
\begin{proof}

We need the following assumption:
we assume that the noise introduced by the imputed reward is heteroscedastic. 
 Formally, let $\rho : X \rightarrow R$
 be a continuous, positive function, such that $\epsilon_t$ is conditionally
 $\rho(x_t)$-subgaussian, that is for all $t \geq 1$ and $\rho_t = \rho(x_t)$,
\begin{equation}
\forall\; \lambda \in R, \quad E[e^{\lambda n_t} | F_{t-1}, x_t] \leq \exp\left(\frac{\lambda^2 \rho_t^2}{2}\right) \text{ .}\label{eq: noise assumption}
\end{equation}
Note that this condition implies that the noise has zero mean, and common examples include Gaussian, Rademacher, and uniform random variables
We need the following lemma, 

\begin{lem} \label{lem:ct} 
Assuming that, the measurement noise $\epsilon_t$ issatisfies  assumption \eqref{eq: noise assumption}.
With probability $1-\delta$, where $0 < \delta < 1$ and  $\theta^*$ lies in the confidence ellipsoid.
\begin{eqnarray*}
C_{t}=\{ \theta: \|\theta-\hat{\theta}_{t}\|_{A_{t}} \leq c_{t} :=\nonumber \\  \rho_t \sqrt{D \log \frac{\det(\mathbf{A}_{t})^{1/2} \det(\mathbf{S}_t)^{-1/2}}{\delta}}+ \|\theta^*\|_2\}
\end{eqnarray*}
\end{lem}

The lemma is adopted from theorem 2 in \cite{abbasi2011improved} using the noise being heteroscedastic. We follow the same step of proof, the main difference is that they have $\mathbf{A}_T=\lambda \mathcal{I}_D+ \sum_{t=1}^T x_tx_t^\top$ and we have $\mathbf{A}_T=\mathbf{S}_T+ \sum_{t \not\in s} x_tx_t^\top $ with $\mathbf{S}_T=\mathcal{I}_D+ \sum_{t \in s} x_tx_t^\top $ with $s \subset T$ contains the contexts with missing rewards .
 
$ R(t) = [ x_t^{*\top} \theta^*-x_t^\top \theta_t] = [ x_t^{*\top} \theta^*- x_t^\top \theta^l_t ]+[x_t^\top \theta_t^l -x_t^\top \theta_t]$ 

where $ \theta^l$ is the parameter of the classical LINUCB, and then

$ R(t) \leq \| x_t^{*\top} \theta^*- x_t^\top \theta^l_t \|_2+\|x_t^\top \theta^l_t -x_t^\top \theta_t\|_2$ 
 
Now we investigate $\| x_t^{*\top} \theta^*- x_t^\top \theta_t^l \|_2$ and $\|x_t^\top \theta_t^l -x_t^\top \theta_t\|_2$ separately.

Following the same step as the proof of theorem 2 in \cite{abbasi2011improved} we also have the following,

  




$\| x_t^{*\top} \theta^*- x_t^\top \theta_t^l \|_2\leq 2 c_t \|x_t\|_{\mathbf{A}_{t}^{-1}}$, and  using Cauchy-Schwarz with $\|\theta_t^l -\theta_t\|_2\leq \epsilon_t $, we get

$\|x_t^\top \theta^l_t -x_t^\top \theta_t\|_2 \leq \epsilon_t  \|x_t\|_{\mathbf{A}_{t}^{-1}} $ and then,

$R(t) \leq (2 c_t+\epsilon_t) \|x_t\|_{\mathbf{A}_{t}^{-1}}$

Since $x^{\top}\theta_{t}^* \in [-1,1]$ for all $x \in X_t $ then we have $R(t) \leq 2$. Therefore,

$R(t) \leq \text{min}\{(2 c_t+\epsilon_t)\|x\|_{\mathbf{A}^{-1}_{t}},2\} \leq 2( c_t+\epsilon_t/2) \; \text{min}\{\|x\|_{\mathbf{A}^{-1}_{t}},1\}$

Our bound on the imputed reward assures $\epsilon_t \leq c_t$. Therefore,

$[R(t)]^2 \leq 9 c_t^2 \text{min}\{\|x\|^2_{\mathbf{A}^{-1}_{t}},1\}$

we have,

$R(T) \leq \sqrt{T\sum_{t=1}^{T}[R(t)]^2}= \sqrt{  \sum_{t=1}^T 9 c_t^2  \text{min}\{\|x\|^{2}_{\mathbf{A}^{-1}_{t}},1\}}$ 

$R(T)\leq 3 c_T \sqrt{ T} \sqrt{ \sum_{t=1}^T   \text{min}\{\|x\|^{2}_{\mathbf{A}^{-1}_{t}},1\}}$, with $c_{T}$ monotonically increasing

since $x \leq 2\,\log(1+x)$ for $x \in [0,1]$, 

we have $\sum_{t=1}^{T} \text{min}\{\|x_t\|^2_{\mathbf{A}_{t}^{-1}}, 1\} \leq 2 \sum_{t=1}^{T} \log(1+\|x_t\|^2_{\mathbf{A}^{-1}_t})\leq 2 (\log\det(\mathbf{A}_{T})-\log\det(\mathbf{S}_T))$, 

here we also use the fact that we have $\mathbf{A}_T=\mathbf{S}_T+ \sum_{s=1}^T x_sx_s^\top $  to get the last inequality.  

$R(T)\leq 3 c_T \sqrt{2(\log\det(\mathbf{A}_{T})-\log\det(\mathbf{S}_T))}$
 
by upper bounding $c_{T}$ using lemma \ref{lem:ct} we get our result.
\end{proof}


\section{Experiments}

In order to verify the proposed MLINUCB methodology, we ran the LINUCB and MLINUCB algorithms on four different datasets, three derived from the UCI Machine Learning Repository \footnote{https://archive.ics.uci.edu/ml/datasets.html}: Covertype, CNAE-9, and Internet Advertisements, and one external dataset : Warfarin. The Warfarin dataset concerns the dosage of the drug Warfarin, where each record consists of a context of patient information and the corresponding appropriate dosage or action. The reward is then defined as 1 if the correct action is chosen and 0 otherwise. The details for each of these datasets are summarized in the Table \ref{table:Synthetic}. 

\begin{table}[ht]
	\centering
	\caption{Datasets}
	\resizebox{0.6\columnwidth}{!}{
		\begin{tabular}{ l | c | r | r }
			Datasets                  & Instances    & Features   & Classes \\ \hline
            Covertype                & 500 000      & 95           &  7\\
            CNAE-9                   & 1080         & 856          &  9\\
            Internet Advertisements  & 3279         & 1558         &  2\\
            Warfarin				 & 5528         & 93           & 3 \\
			
		\end{tabular}
	}
	\label{table:Synthetic}
\end{table}
To evaluate the performance of MLINUCB and LINUCB we utilize an accuracy metric that checks the equality of the selected action and the best action, which is revealed for the purposes of evaluation. Defined as such, accuracy is inversely proportional to regret. In the following experiments we fix $m=1, \alpha=0.25$ and utilize the mini batch K-means algorithm for clustering. In Table \ref{accuracy}, we report the total average accuracies of running LINUCB and MLINUCB with 2, 5, 10, 15, and 20 clusters on each dataset.

\begin{table}[ht]
\centering
\caption{Total average accuracy}
\label{accuracy}
\begin{tabular}{l|l|l|l|l}
                       \multicolumn{5}{c}{10\% Missing Rewards}                                                           \\ \hline
                      & Covertype      & CNAE-9         & Internet Ads      & Warfarin       \\ \hline
LINUCB                  & \textbf{0.884}          & 0.644          & 0.866                        & 0.643          \\ \hline
MLINUCB - $N=2$         & 0.869          & 0.643          & 0.898                        & 0.643                   \\ 
MLINUCB - $N=5$         & 0.874          & 0.626          & 0.895                        & \textbf{0.656}                    \\ 
MLINUCB - $N=10$        & 0.880          & 0.664          & 0.894                        & 0.650                    \\ 
MLINUCB - $N=15$        & 0.877          & \textbf{0.678} & \textbf{0.902}               & 0.647                    \\ 
MLINUCB - $N=20$        & 0.878          & 0.675          & 0.898                        & 0.653                    \\ \hline
\end{tabular}
\\\vspace{\baselineskip}
\begin{tabular}{l|l|l|l|l}
                       \multicolumn{5}{c}{50\% Missing Rewards}                                                           \\ \hline
                      & Covertype      & CNAE-9         & Internet Ads & Warfarin       \\ \hline
LINUCB                & \textbf{0.884} & 0.566          & 0.824                   & 0.615          \\ \hline
MLINUCB - $N=2$       & 0.838          & 0.578          & 0.888                   & 0.630             \\ 
MLINUCB - $N=5$       & 0.847 		   & 0.546          & 0.896                   & \textbf{0.641}         \\ 
MLINUCB - $N=10$      & 0.863          & 0.592          & 0.897                   & 0.640          \\ 
MLINUCB - $N=15$      & 0.854          & \textbf{0.608} & \textbf{0.903}          & 0.638             \\ 
MLINUCB - $N=20$      & 0.853          & 0.592          & 0.901                   & 0.639              \\ \hline
\end{tabular}
\\\vspace{\baselineskip} 
\begin{tabular}{l|l|l|l|l}

                       \multicolumn{5}{c}{75\% Missing Rewards}                                                           \\ \hline
                      & Covertype      & CNAE-9         & Internet Ads & Warfarin       \\ \hline
LINUCB                & \textbf{0.880} & 0.483          & 0.786                   & 0.610          \\ \hline
MLINUCB - $N=2$       & 0.784          & 0.461          & 0.881                   & 0.594             \\ 
MLINUCB - $N=5$       & 0.797          & 0.494          & 0.890                   & 0.612             \\ 
MLINUCB - $N=10$      & 0.837          & \textbf{0.521} & 0.887                   & \textbf{0.624}         \\ 
MLINUCB - $N=15$      & 0.824          & 0.500          & 0.891                   & 0.600          \\ 
MLINUCB - $N=20$      & 0.819          & 0.493          & \textbf{0.896}          & 0.611  \\ \hline 
\end{tabular}
\vspace{\baselineskip}
\end{table}

As the MLINUCB regret upper bound is lower than the LINUCB regret upper bound when $\epsilon$ is small, minimizing clustering error is critical to performance. Accordingly, successful MLINUCB operates on the assumption that the context vectors live in a manifold that can be described by a set of clusters. Thus MLINUCB has the potential to outperform LINUCB when this manifold assumption holds, specifically when the number of clusters chosen adequately describes the structure of the context vector space. Visualizing the context vectors suggests that some of our test datasets violate this assumption, some respect this assumption, and when an appropriate number of clusters is chosen, MLINUCB performance aligns as expected. 

\begin{figure}[H]
\centering
\subfigure[Context vector visualization with 5 clusters and 2D PCA. 2D PCA captures 13.9\% of the variance in the CNAE-9 dataset.]{\includegraphics[scale=0.19]{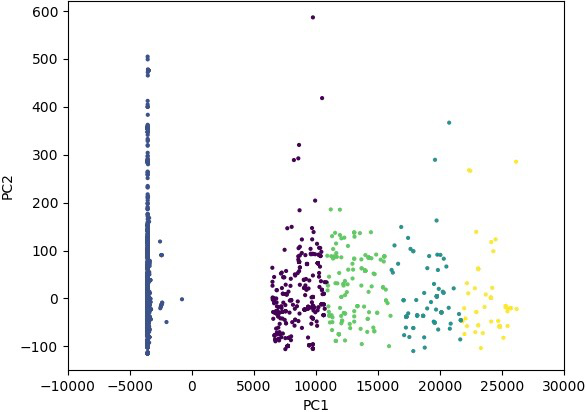}}
\subfigure[LINUCB  and MLINUCB accuracy comparison]{\includegraphics[scale=0.1]{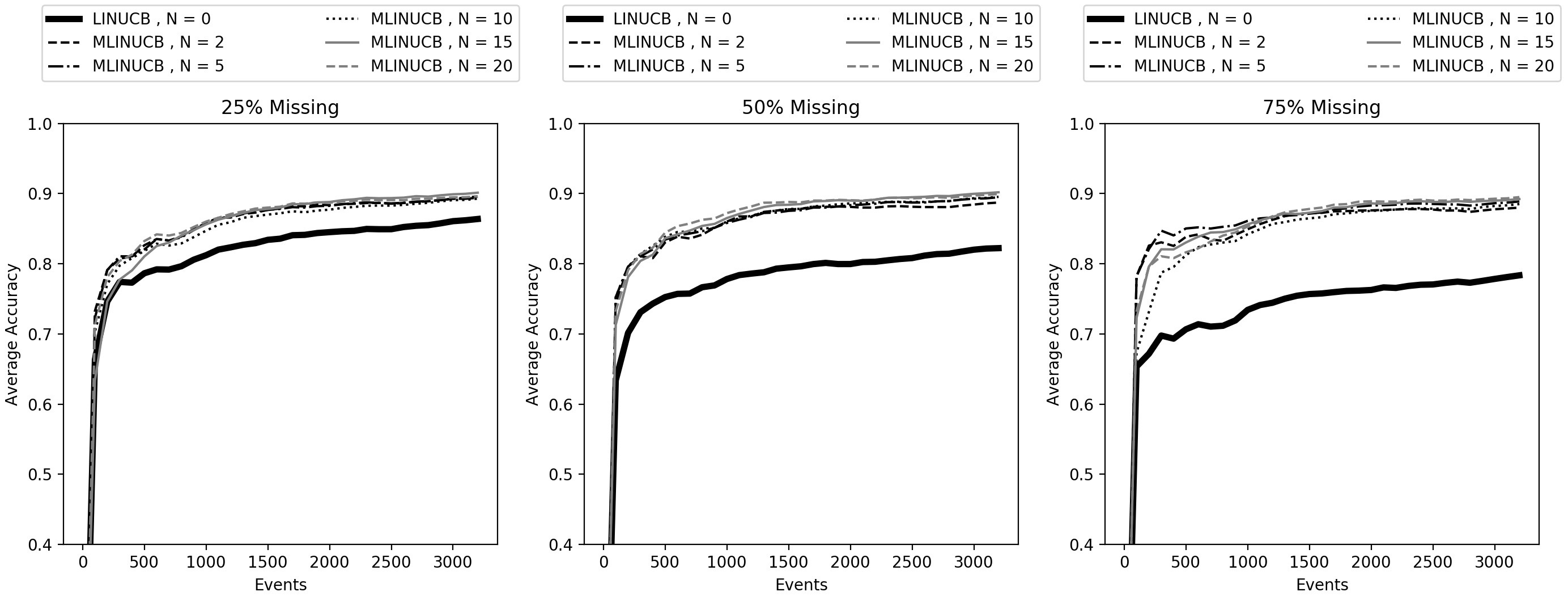}}
\caption{Internet Advertisements}
\label{Advertisement_Acc}
\end{figure}

\begin{figure}[H]
\centering
\subfigure[Context vector visualization with 5 clusters and 2D PCA. 2D PCA captures 98.2\% of the variance in the Warfarin dataset.]{\includegraphics[scale=0.2]{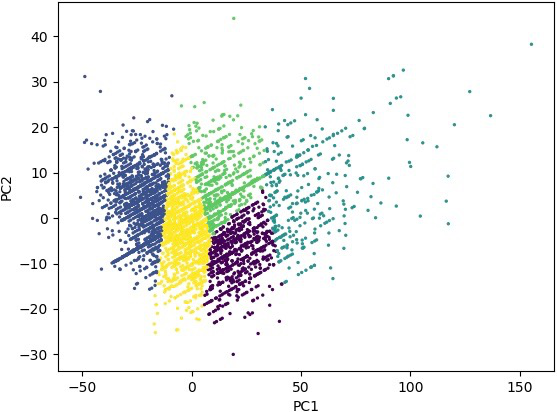}}
\subfigure[LINUCB  and MLINUCB accuracy comparison]{\includegraphics[scale=0.1]{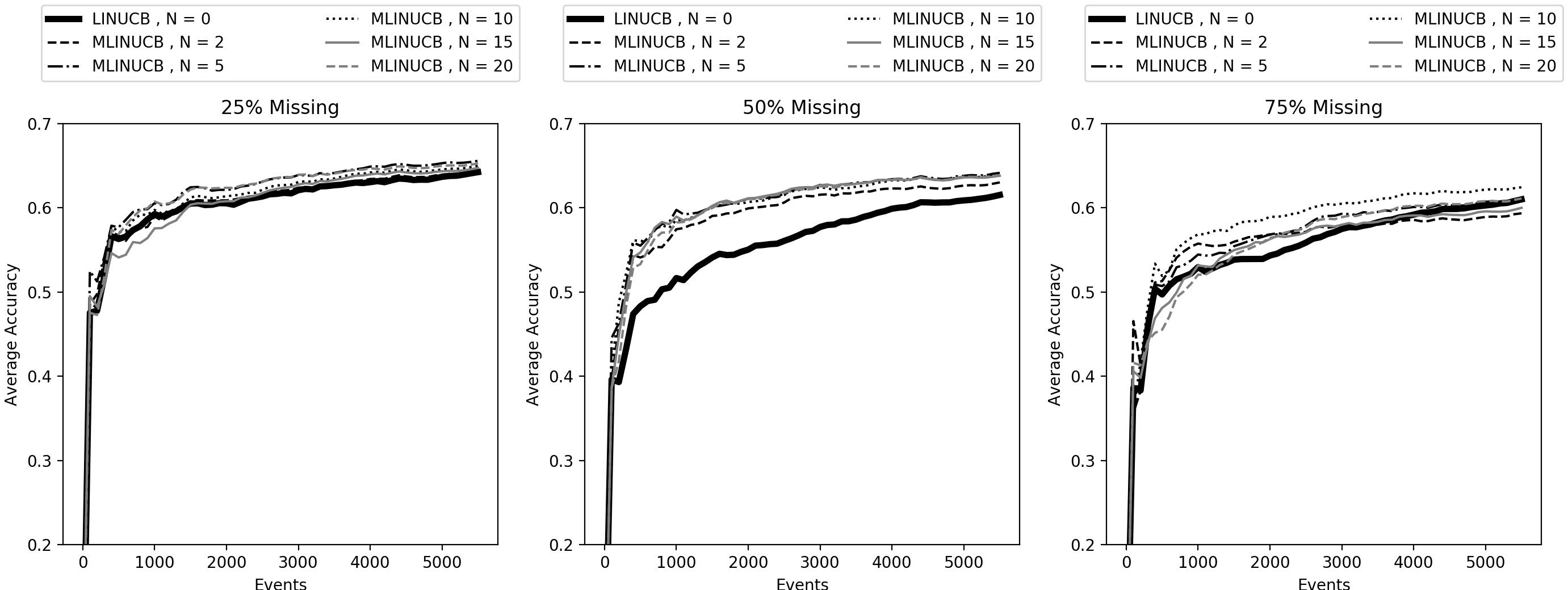}}
\caption{Warfarin}
\label{warfarin_acc}
\end{figure}

Consider the Internet Advertisements and Warfarin datasets, where 2D projections of the context vectors capture the majority of the variance in the context vector space, 100.0\% and 98.2\% respectively. In Figures \ref{Advertisement_Acc} and \ref{warfarin_acc} the projected context vector spaces appear clustered, not randomly scattered, and MLINUCB outperforms LINUCB for most choices of $N$, the number of clusters. The Internet Advertisements dataset yields the best results - when switching from LINUCB to MLINUCB algorithms, accuracy jumps from $86.6\%$ to $90.2\%$ when $25\%$ of the reward data is missing, from $82.4\%$ to $90.3\%$ when $50\%$ of the reward data is missing, and from $78.6\%$ to $89.6\%$ when $75\%$ of the reward data is missing.

\begin{figure}[H]
\centering
\subfigure[Context vector visualization with 5 clusters and 2D PCA. 2D PCA captures 29.7\% of the variance in the Covertype dataset.]{\includegraphics[scale=0.2]{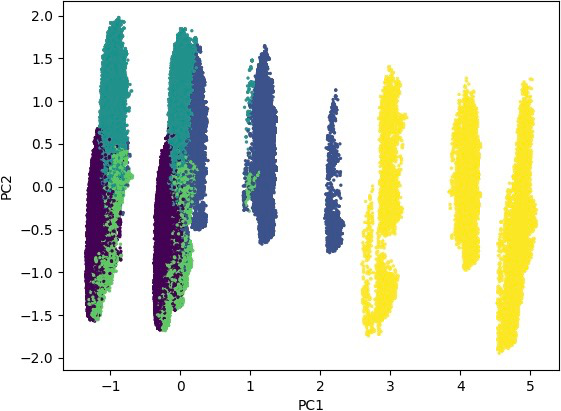}}
\subfigure[LINUCB  and MLINUCB accuracy comparison]{\includegraphics[scale=0.1]{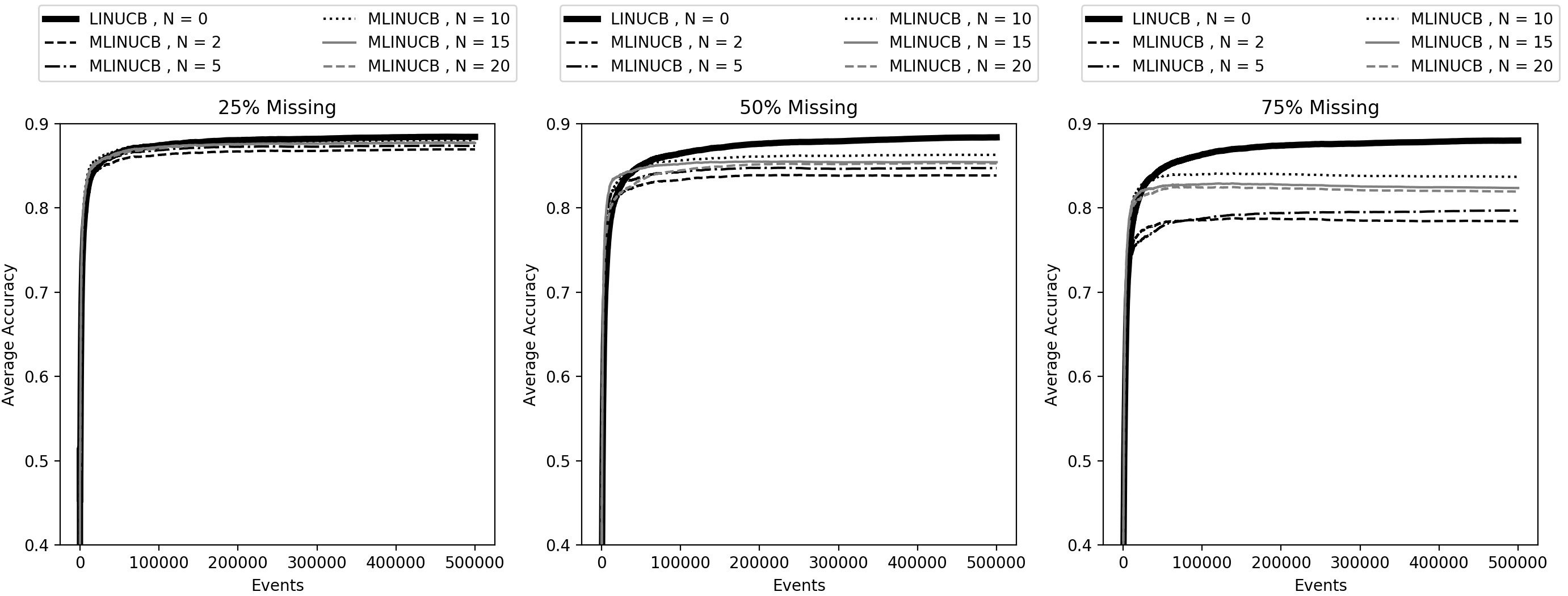}}
\caption{Covertype}
\label{covtree1_acc}
\end{figure}

\begin{figure}[H]
\centering
\subfigure[Context vector visualization with 5 clusters and 2D PCA. 2D PCA captures 13.9\% of the variance in the CNAE-9 dataset.]{\includegraphics[scale=0.2]{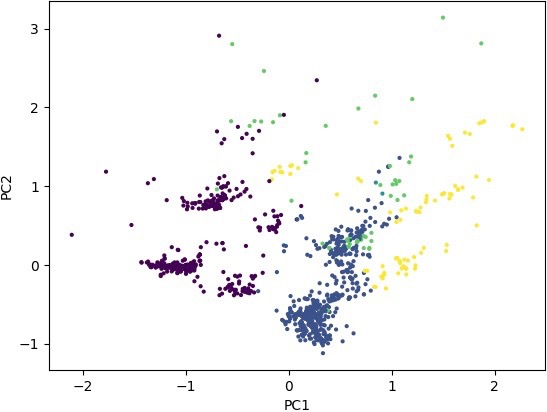}}
\subfigure[LINUCB  and MLINUCB accuracy comparison]{\includegraphics[scale=0.1]{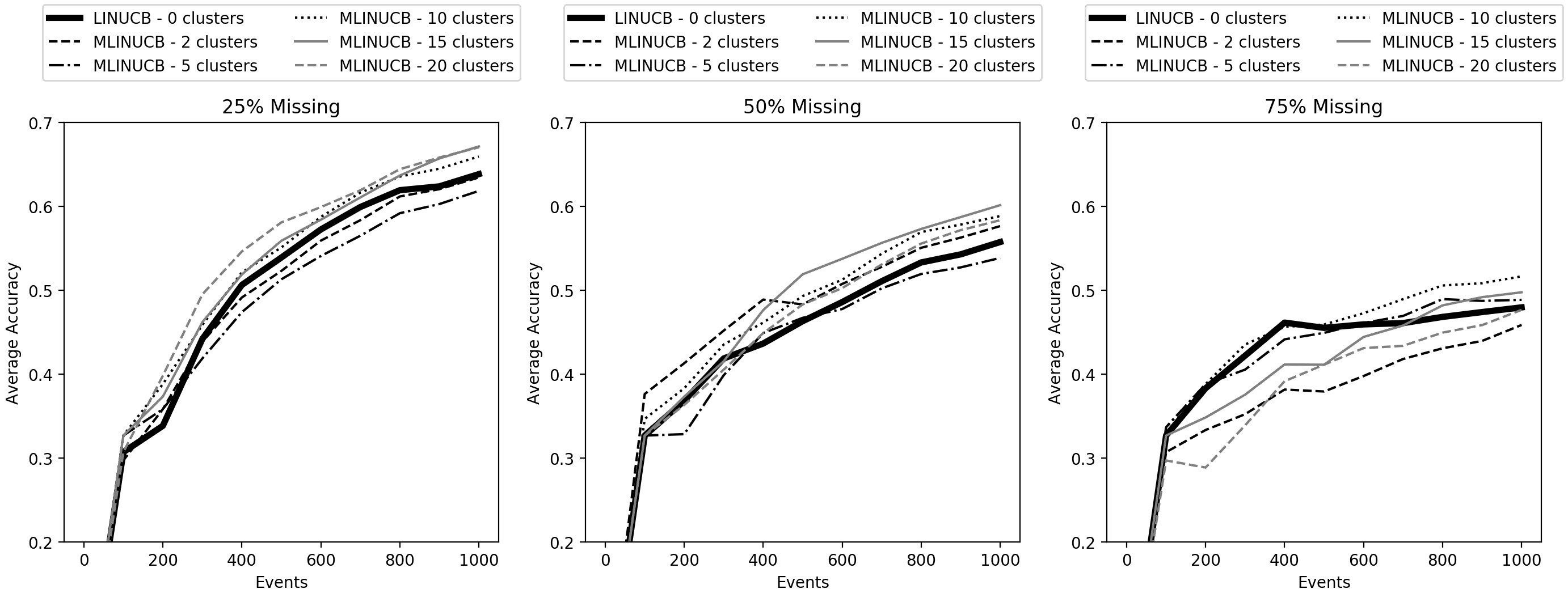}}
\caption{CNAE-9}
\label{CNAE_acc}
\end{figure}

Although the 2D projections of the Covertype and CNAE-9 context vectors in Figures \ref{covtree1_acc} and \ref{CNAE_acc} appear well clustered, both projections only capture a small amount of the variance in the context vector space, $29.7\%$ in the Covertype dataset and $13.9\%$ in the CNAE-9 dataset. MLINUCB results do not show improvement for the cases tried for Covertype dataset suggesting that the Covertype dataset violates the manifold assumption for the context space. However in the CNAE-9 dataset, we see that MLINUCB outperforms LINUCB for most choices of $N$, which supports the observation that the context space is clustered. 


\begin{figure}[h]
\label{CNAE9_alpha}
\centering
{\includegraphics[scale=0.11]{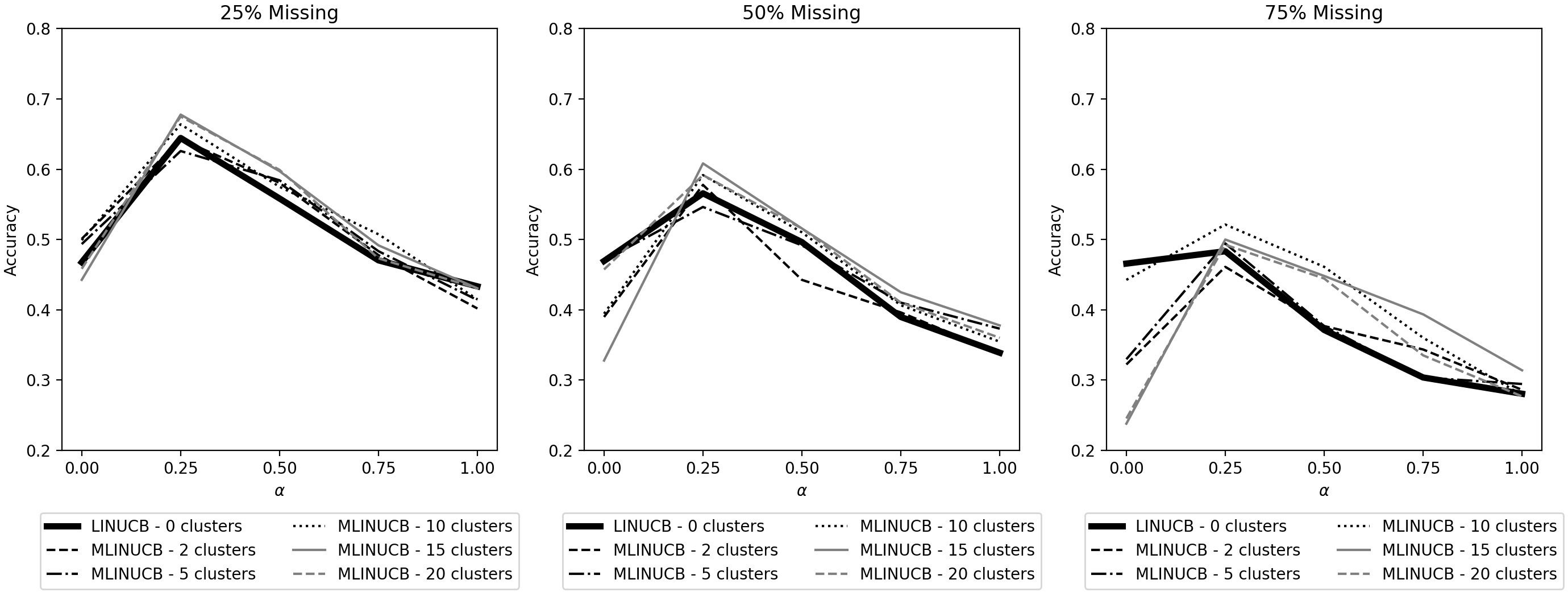}}
\caption{LINUCB and MLINUCB cumulative accuracies on CNAE-9 across various $\alpha$}
\end{figure}

Taking a more in depth look at the CNAE-9 dataset, in Figure \ref{CNAE9_alpha}, we vary LINUCB and MLINUCB's common hyperparameter $\alpha$, which controls the ratio of exploration to exploitation, and see that MLINUCB continues to result in higher accuracies than LINUCB for most $\alpha$. 

Note that $N$, the number of clusters, is a hyperparameter of the algorithm and while initialized \emph{a-priori}, it could be changed and optimized online as more context vectors are revealed. Alternatively, we could leverage clustering algorithms that do not initialize $N$ \emph{a-priori} and learn the best $N$ from the available data.

\section{Conclusions and Future Work}
In this paper we studied the effect of data imputation in the case of missing rewards for multi-arm bandit problems. We prove an upper bound for the total regret in our algorithm following the CBP upper bound. Our MLINUCB algorithm shows improvements over LINUCB in terms of total average accuracy for most cases. The main observation here is that when the context vector space lives in a clustered manifold, we can take advantage of this structure and impute the missing reward at each step given similar context in previous events. 
A very obvious next step is to try using the weighted average introduced in equation \ref{waverage} with $m$ greater than $1$. This would use more topological information from the context feature space and wouldn't rely on a single cluster. Additionally, the algorithm doesn't rely on a fixed value for $N$ so we could optimize the value of $N$ at each event using some clustering metric to find the best $N$ at each time. This work can also be extended by replacing the simple clustering step with more complex methodologies to learn a representation of the context vector space, for example sparse dictionary learning.

\bibliography{references}
\bibliographystyle{plain}


\end{document}